# Tutor Move Taxonomy: A Theory-Aligned Framework for Analyzing Instructional Moves in Tutoring

Zhuqian Zhou, Kirk Vanacore, Tamisha Thompson, Jennifer St John, René Kizilcec

National Tutoring Observatory
Cornell University

**Abstract**. Understanding what makes tutoring effective requires methods for systematically analyzing tutors' instructional actions during learning interactions. This paper presents a tutor move taxonomy designed to support large-scale analysis of tutoring dialogue within the National Tutoring Observatory. The taxonomy provides a structured annotation framework for labeling tutors' instructional moves during one-on-one tutoring sessions. We developed the taxonomy through a hybrid deductive–inductive process. First, we synthesized research from cognitive science, the learning sciences, classroom discourse analysis, and intelligent tutoring systems to construct a preliminary framework of tutoring moves. We then refined the taxonomy through iterative coding of authentic tutoring transcripts conducted by expert annotators with extensive instructional and qualitative research experience. The resulting taxonomy organizes tutoring behaviors into four categories: tutoring support, learning support, social-emotional and motivational support, and logistical support. Learning support moves are further organized along a spectrum of student engagement, distinguishing between moves that elicit student reasoning and those that provide direct explanation or answers. By defining tutoring dialogue in terms of discrete instructional actions, the taxonomy enables scalable annotation, computational modeling of tutoring strategies, and empirical analysis of how tutoring behaviors relate to learning outcomes.

# 1. Introduction

The National Tutoring Observatory is building a large-scale repository of tutoring data spanning diverse instructional contexts, from extended, high-dosage sessions to on-demand tutoring for targeted academic support. The central aim of this effort is to understand what makes tutoring effective by linking tutor moves—and patterns of tutor moves—to student learning outcomes. A core component of this repository is a structured annotation system capturing tutors' instructional actions (i.e., tutor moves), enabling systematic analysis of pedagogical practice at scale.

To develop this tutoring taxonomy, we adopted a hybrid inductive/deductive approach. First, we conducted an extensive review of relevant instructional, tutoring, discourse, and learning science literature. A brief overview of this review is provided below. We then used that construct from the literature to develop a preliminary taxonomy. Our next step is to refine this codebook based on the tutoring transcripts and expert evaluation. Presently, we are working with two expert annotators with over thirty years of experience in mathematics instruction and instructional coaching, doctorates in the learning sciences, and experience in qualitative coding. Both annotators currently work in public school settings. They have been reviewing tutoring data from multiple providers and iteratively refining the taxonomy based on recurring patterns and contextual nuances. The draft taxonomy presented below reflects the current synthesis of this literature-informed and expert-driven development process.

Our evaluation framework is intentionally inductive—emerging from the data: rather than presuming what constitutes "good" tutoring, we analyze tutors' observed behaviors and empirically examine how different move patterns relate to student outcomes. Complementing this qualitative and descriptive work, we apply quasi-experimental machine learning methods—such as leave-one-out potential outcomes and causal forests[1]—to estimate heterogeneous session effectiveness. These causal and predictive analyses allow us to identify which tutoring behaviors and interaction patterns are associated with stronger learning gains, advancing a data-driven understanding of what drives successful tutoring experiences.

## 2. Deductive & Inductive Taxonomy Development

## 2. 1 Theoretical Background used for Deductive Taxonomy Development

We grounded the tutoring taxonomy in classic work from cognitive science, the learning sciences, and tutoring research that conceptualizes tutoring as a sequence of goal-directed instructional moves—diagnosing understanding, scaffolding, explaining, prompting, giving feedback, and supporting motivation and affect (Wood et al., 1976; Lesh et al., 1987; Merrill et al., 1995; Chi et al., 2001; VanLehn et al., 2003; Chi et al., 2008; Renkl et al., 2009; Van de Pol et al., 2010; Renkl, 2011; Rittle-Johnson et al., 2015; Sweller et al., 2019; Ainsworth, 2006). This work connects closely to formative assessment and feedback research, emphasizing how instructors elicit, interpret, and respond to student thinking (Black & Wiliam, 1998, 2009; Hattie & Timperley, 2007; Gotwals & Birmingham, 2016; Wisniewski et al., 2020). We also reviewed intelligent tutoring systems (ITS) research that formalizes tutoring strategies as discrete policy actions—such as hints, prompts, adaptive questions, and feedback—across cognitive tutors and natural-language dialogue systems (Graesser et al., 1995; Koedinger & Corbett, 2006; Koedinger & Aleven, 2007; Ohlsson et al., 2007; Kim et al., 2006; Olney et al., 2010; Rus et al., 2015; Vail & Boyer, 2014).

We further drew on classroom observation frameworks and discourse-analytic traditions that operationalize instructional practice into codable behaviors and talk moves. This includes large-scale observation and evaluation systems such as CLASS, the Framework for Teaching, MQI, PLATO, QST, and UTOP, as well as evidence from the MET project demonstrating the reliability and predictive validity of structured rubrics (Pianta et al., 2008; Danielson, 2014; Grossman et al., 2013; Learning Mathematics for Teaching Project, 2011; Schultz & Pecheone, 2015; Walkington & Marder, 2018; Kane & Staiger, 2012). Within classroom discourse research, we reviewed

---

[1] Given a high-dimensional matrix of covariates, these methods can estimate near-individual-level treatment effects. We are using them in an ongoing study of human tutoring support in an intelligent tutoring intervention (preprint forthcoming)

Accountable Talk, Talk Science, and related linguistic and interactional coding frameworks that specify fine-grained instructional and conversational moves—such as revoicing, pressing for reasoning, and coordinating participation—alongside STEM- and design-focused episode-level coding schemes (Sinclair & Coulthard, 1992; Wolf et al., 2005; Chapin et al., 2009; Michaels & O'Connor, 2012; Atman et al., 2007; Menekse et al., 2019; Hurst et al., 2023).

Finally, we incorporated emerging NLP, machine learning, and educational data science literature that treats instructional moves as labeled units for scalable modeling and automated analysis. This includes annotated dialogue corpora such as TalkMoves, uptake datasets, and accountable-talk detectors, as well as shared tasks and benchmarks for AI teacher response generation and student dialogue understanding (Demszky et al., 2021; Suresh et al., 2022; Kupor et al., 2023; Tack & Piech, 2022; Tack et al., 2023; Mim et al., 2025). These efforts conceptualize tutoring as a sequence of process tokens—categories like probing student thinking, hinting, or revoicing—that serve as units of analysis for both humans and machines (Roscoe & Chi, 2007; Boyer et al., 2010). Reviewing these resources highlighted a core design gap: existing schemes are often domain-specific, optimized either for human observation or narrow modeling tasks, or mismatched in grain size, motivating the need for a domain-general tutoring moves taxonomy that is simultaneously theory-aligned and AI-ready (Kim et al., 2006; Olney et al., 2010; Rus et al., 2015; Lin et al., 2022).

## 2.2 Expert Annotation for Inductive Taxonomy Refinement

Following the development of the preliminary literature-informed taxonomy, we conducted an inductive coding process to refine and validate the codebook using authentic tutoring transcripts. The goal of this process was to ensure that the taxonomy captured the range of instructional actions observed in real tutoring interactions while maintaining alignment with the theoretical constructs described above.

Two expert annotators—authors three and four—participated in the coding process. Both are veteran teachers with doctoral training in the learning sciences and extensive experience in qualitative research and instructional practice. The annotators reviewed tutoring sessions from multiple providers and applied the preliminary coding scheme to segments of tutor–student dialogue.

The inductive development process involved multiple iterative rounds of coding and discussion organized into two phases: taxonomy development and refinement with inter-rater reliability (IRR) assessment.

### *2.3 Phase 1: Taxonomy Development.*
In the first phase, the annotators conducted coding sessions both collaboratively and independently, revising the codebook whenever definitions, distinctions, or category boundaries proved unclear. Through repeated examination of tutoring transcripts, the annotators proposed additions, merges, and refinements to the taxonomy to better capture recurring instructional patterns. These sessions focused on identifying ambiguities in the preliminary coding scheme and improving the clarity and scope of the code definitions. After several rounds of collaborative and independent coding (approximately 10 hours), the codebook reached a sufficiently stable form to proceed to reliability testing.

### *2.4 Phase 2: Codebook Refinement and Inter-Rater Reliability.*
In the second phase, the annotators coded tutoring sessions independently using the updated codebook. They then met to review discrepancies and examine instances of disagreement, ambiguous cases, or interaction patterns not fully captured by the existing taxonomy. Disagreements were resolved through structured discussions and social moderation, during which the annotators jointly reviewed representative excerpts and refined operational definitions, category boundaries, and decision rules.

Through these repeated cycles of independent coding and moderated refinement, the codebook gradually stabilized, resulting in clearer definitions, improved distinctions between closely related tutoring moves, and the addition or consolidation of categories where necessary. In the first round of coding, the annotators achieved moderate IRR (Cohen's $\kappa$ = .65), which improved in the second round (Cohen's $\kappa$ = .78) after clarifications and minor alterations to the taxonomy. The resulting taxonomy reflects both theoretical grounding and inductive insights derived from observed tutoring practice and is designed to capture the range of instructional actions that occur in one-on-one tutoring contexts.

## 3. Tutor Move Taxonomy

Table 1 presents the current draft of our tutor move taxonomy. This taxonomy is intended to function as a living framework that will evolve as the National Tutoring Observatory ingests additional data across tutoring contexts. The current taxonomy focuses on one-on-one tutoring support, but will likely expand to constructs more relevant to small group tutoring as we ingest more data. As new instructional patterns, institutional practices, and edge cases emerge, the taxonomy will be iteratively refined to better capture the full range of tutoring behaviors and to maintain alignment with both learning theory and empirical evidence.

The current taxonomy is organized into three primary categories. *Tutoring support* moves focus on information gathering and decision-making about how to proceed in the tutoring process. We define probing in this category as brief questions that often invite binary or low-effort responses (e.g., "Did you learn this?" or "Are you with me?"). These moves may carry institutional or procedural value—such as a tutor verbalizing plans for approaching a problem—but they are oriented more toward the tutor's planning and regulation than toward advancing student learning directly. *Learning support* moves represent the instructional core of most tutoring sessions. These moves are placed on a spectrum ranging from those designed to elicit active student thinking, explanation, and engagement to those in which the student is a more passive recipient of information or worked examples. This category captures the pedagogical actions most directly tied to conceptual understanding, procedural fluency, and transfer. Notably, the taxonomy distinguishes between "probing", defined as asking a categorical or yes–no question, and "prompting", in which the tutor asks an open-ended question. *Social and emotional support* moves encompass behaviors through which tutors encourage students, build rapport, and support motivation, confidence, and persistence. In addition, we code logistical support moves that address session management, technology, scheduling, or other operational needs that shape the tutoring experience.

**Table 1. NTO Tutor Move Taxonomy.**

| Category | Code | Definition |
|---|---|---|
| TUTORING SUPPORT *Moves for supporting tutors' tutoring decision-making* | ASKING_TO_CLARIFY_CONTEXT | Tutor asks the student about the problem to understand the context. |
| | PROBING_PRIOR_KNOWLEDGE | Tutor probes about student's prior knowledge |
| | PROBING_UNDERSTAND | Tutor questions whether student understood the statement made or step(s) done. |
| | CORRECTING_OWN_ERROR | Tutor explicitly recognizes a mathematical error they made. |
| | STRATEGIZING | Tutor discusses plans with the student to complete the problem. |
| *Higher student engagement* ↑ LEARNING SUPPORT Spectrum ↓ *Lower student engagement* | PROMPTING_RELATED_CONCEPTS | Tutor prompts the student to make connections to related concepts without giving detailed explanations of the related concepts. |
| | PROMPTING_ALTERNATIVE_REPRESENTATION | Tutor asks the student to represent the idea in a different form. [Examples: model from word problem, diagram, table, equation, number line, words… This will belong to the example column.] |
| | PROMPTING_SELF_EXPLANATION | Tutor prompts the student to explain actions or knowledge. |
| | PROMPTING_NEXT_STEP | Tutor prompts the student to produce any step without adding new information. |
| | PROMPTING_SELF_CORRECTION | Tutor prompts the student to correct their own mistakes without giving away the right way. |
| | FEEDBACK_CORRECT | Tutor indicates that the student has provided a correct response. |
| | FEEDBACK_INCORRECT | Tutor indicates that the student has provided an incorrect response. |
| | FEEDBACK_NEUTRAL | Tutor comments without confirming accuracy or inaccuracy. |
| | REVOICING | Tutor says what the student said in different ways. |
| | RESTATING | Tutor says what the student said verbatim. |
| | GIVING_HINT | Tutor provides a suggestion to guide the student toward a solution without revealing the answer. |
| | GIVING_EXAMPLE | Tutor uses an example that is different from the student's problem. |
| | EXPLAINING_CONCEPTUAL | Tutor gives instruction on concepts or facts. |
| | EXPLAINING_PROCEDURAL | Tutor gives instruction about the algorithmic steps to solve the problem. |
| | GIVING_ANSWER | Tutor provides the solution without including the solution steps. |
| SOCIAL-EMOTIONAL & MOTIVATIONAL SUPPORT *Moves for supporting students' social-emotional well-being and motivation.* | BUILDING_RAPPORT | Tutor engages in relationship-building conversation, including when the tutor humanizes themselves. |
| | ENCOURAGING | Tutor makes encouraging statements to students. |
| | ASKING_FEELING | Tutor asks for the student's feelings. |
| | VALIDATING_FEELING | Tutor validates the student's feelings. |
| | PRAISING_PROCESS | Tutor praises the student's process of solving the problem. |
| | PRAISING_OUTCOME | Tutor praises the student's outcome of student work. |

|  | PRAISING_TRAITS | Tutor praises the student's characteristic. |
| --- | --- | --- |
| LOGISTICAL SUPPORT *Moves dealing with logistical details.* | TECHNICAL_ISSUES | Tutor makes a comment related to technical problems of the tutoring context. |

# Conclusion and Future Directions

This paper introduces a tutor move taxonomy designed to systematically capture the instructional actions that occur during one-on-one tutoring interactions. Drawing on theory from the learning sciences, tutoring research, and classroom discourse analysis, and refined through iterative inductive coding of authentic tutoring transcripts, the taxonomy provides a structured framework for annotating tutoring dialogue. By organizing tutoring behaviors into categories that distinguish between instructional support, student engagement strategies, social-emotional support, and logistical coordination, the framework enables researchers to analyze tutoring interactions in a consistent and theory-informed manner.

Beyond supporting qualitative analysis, the taxonomy is designed to function as a scalable representation of tutoring dialogue that can be used in large-scale computational studies of tutoring practice. Because tutor moves are defined as discrete interaction units, they can serve as labels for supervised learning models and benchmarks for automated analysis of tutoring transcripts. This creates opportunities to link patterns of instructional actions to student outcomes across large datasets, enabling more precise empirical investigations of what makes tutoring effective.

Future work will focus on expanding the taxonomy and scaling the annotation process through AI-assisted methods. In particular, we plan to explore the use of large language models to automatically classify tutor moves, using human annotations as training and validation data. Such systems could substantially accelerate the analysis of large tutoring corpora while maintaining alignment with theory-driven coding schemes. In addition, automated annotation could enable real-time analytics for tutoring systems and professional development tools that provide feedback on instructional practices. By combining expert-informed taxonomies with AI-assisted annotation pipelines, this work aims to support a new generation of research and tools for understanding and improving tutoring at scale.